%% file: main.tex
\documentclass{article}

\usepackage{arxiv}

\usepackage{amsmath,amssymb}
\usepackage{graphicx}
\usepackage{wrapfig}
\usepackage{array}
\usepackage[algoruled,linesnumbered,algo2e,vlined]{algorithm2e}
\usepackage{amssymb}
\usepackage{url}
\usepackage{multirow}
\usepackage{cite}

\usepackage{algorithm}
\usepackage{algorithmic}

\usepackage{comment}

\title{LCP-dropout: Compression-based Multiple Subword Segmentation for Neural Machine Translation}

\author{
Keita Nonaka
\And
Kazutaka Yamanouchi
\And
Tomohiro I
\And
Tsuyoshi Okita
\And
Kazutaka Shimada
\And
Hiroshi Sakamoto\\ \\
Kyushu Institute of Technology, 680-4 Kawazu, Iizuka, Fukuoka 820-8502, Japan \\
\texttt{\{ nonaka.keita331, yamanouchi.kazutaka437  \}@mail.kyutech.jp}\\
  \texttt{ \{tomohiro, okita, shimada, hiroshi\}@ai.kyutech.ac.jp} 
}

\begin{document}
\maketitle
\begin{abstract}
In this study, we propose a simple and effective preprocessing method 
for subword segmentation based on a data compression algorithm.
Compression-based subword segmentation has recently attracted significant attention 
as a preprocessing method for training data in Neural Machine Translation.
Among them, BPE/BPE-dropout is one of the fastest and most effective method compared to conventional approaches.
However, compression-based approach has a drawback in that generating
multiple segmentations is difficult due to the determinism.
To overcome this difficulty, we focus on a probabilistic string algorithm, called locally-consistent parsing (LCP), 
that has been applied to achieve optimum compression.
Employing the probabilistic mechanism of LCP, we propose LCP-dropout for multiple subword segmentation
that improves BPE/BPE-dropout, and
show that it outperforms various baselines in learning from especially small training data. 
\end{abstract}

\input{sec1.tex}

\input{sec2.tex}
\input{sec3.tex}
\input{sec4.tex}
\input{sec5.tex}

\input{sec6.tex}



\end{document}

%% file: sec1.tex
\section{Introduction}

\subsection{Motivation}
Subword segmentation has been established as a standard preprocessing method 
in neural machine translation (NMT)~\cite{Barrault2019,Bojar2018}.
In particular, byte-pair-encoding (BPE)/BPE-dropout~\cite{Sennrich2016,Provilkov2020} is 
the most successful compression-based subword segmentation.
We propose another compression-based algorithm, denoted by LCP-dropout,
that generates multiple subword segmentations for the same input; 
thus, enabling data augmentation especially for small training data.

In NMT, a set of training data is given to the learning algorithm, where
a training data is a pair of sentences from the source and target languages.
The learning algorithm first transforms the given each sentence into a sequence of {\em tokens}.
In many cases, the tokens correspond to words in the unigram language model. 

The extracted words are projected from a high-dimensional space consisting of all words 
to a low-dimensional vector space by {\em word embedding}~\cite{Mikolov2013},  
which enables us to easily handle distances and relationships between words and phrases.
The word embedding have been shown to boost the performance of various tasks~\cite{Socher2013-1,Socher2013-2} 
in natural language processing.
The space of word embedding is defined by a dictionary constructed from the training data,
where each component of the dictionary is called {\em vocabulary}.
Embedding a word means representing it by a set of related vocabularies.

Constructing appropriate dictionary we tackle in this study is one of the most important tasks.
Here, consider a simplest strategy that uses the words themselves in the training data as the vocabularies.
If a word does not exist in the current dictionary, it is called an unknown word, and the algorithm decides
whether or not to register it in the dictionary.
Using a sufficiently large dictionary can reduce the number of unknown words as much as desired;
however, as a trade-off, overtraining is likely to occur, so the number of vocabularies is usually limited to $16\text{k}$ and $32\text{k}$. 
Therefore, the subword segmentation has been widely used to construct a small dictionary with high generalization performance~\cite{Creutz2007,Schuster2012,Chitnis2015,Kunchukuttan2016,Banerjee2018}.

\subsection{Related works}
Subword segmentation is a recursive decomposition of a word into substrings.
For example, let the word `study' be registered as a current vocabulary.
By embedding other words 'studied' and 'studying', we can learn that these three words are similar.
However, each time a new word appears, the number of vocabularies grows monotonically.

On the other hand, when we focus on the common substrings of these words, 
we can obtain a decomposition, such as `stud\_y', `stud\_ied', and `stud\_ying' with the explicit blank symbol `\_'.
Therefore, the idea of subword segmentation is not to register the word itself as a vocabulary 
but to register its subwords.
In this case, `study' and `studied' are regarded as known words 
because they can be represented by combining subwords already registered.
These subwords can also be reused as parts of other words (e.g., student and studied), which can suppress 
the growth of vocabulary size.

In the last decade, various approaches have been proposed along this line.
SentencePiece~\cite{Kudo2018} is a pioneering study based on likelihood estimation 
over unigram language model, having high performance.
Since maximum likelihood estimation requires a quadratic time in the size of training data and the length of longest subword, 
a simpler subword segmentation~\cite{Sennrich2016} based on BPE~\cite{Byte-Pair-Encoding1994,Re-pair2000}, 
which is known as one of fastest data compression algorithms, 
and therefore has many applications, especially in information retrieval~\cite{Shinohara1997,Kida2003} has been proposed.

BPE-based segmentation starts from the state where a sentence is regarded as a sequence of vocabularies
where the set of vocabularies is initially identical to the set of alphabet symbols (e.g., ASCII characters).
BPE calculates the frequency of any bigram,
merges all occurrences of the most frequent bigram, and registers the bigram as a new vocabulary.
This process is repeated until the number of vocabularies reaches the limit.
Thanks to the simplicity of the frequency-based subword segmentation,
BPE runs in linear time in the size of input string.

However, frequency-based approach may generates inconsistent subwords for same substring occurrences.
For example, 'impossible' and its substring 'possible' are possibly decomposed into undesirable subwords,
such as 'po\_ss\_ib\_le' and 'i\_mp\_os\_si\_bl\_e', depending on the frequency of bigrams.
Such merging disagreement can also be caused by misspellings of words or grammatical errors.
BPE-dropout~\cite{Provilkov2020} proposed a robust subword segmentation for this problem by ignoring each merging with a certain probability.
It has been confirmed that BPE-dropout can be trained with higher accuracy than 
the original BPE and SentencePiece on various languages.

\subsection{Our contribution}

We propose LCP-dropout: a novel compression-based subword segmentation 
employing the stochastic compression algorithm, 
called locally-consistent parsing (LCP)~\cite{Cormode2007,Jez2016}, for improving the shortcomings of BPE.
Here, we describe an outline of original LCP.
Suppose we are given an input string and a set of vocabularies, where similarly to BPE,
the set of vocabularies is initially identical to the set of symbols appearing in the string.
LCP randomly assigns the binary label for each vocabulary.
Then, we obtains a binary string corresponding to the input string where the bigram `10' works as a landmark.
LCP merges any bigram in the input string corresponding to a landmark in the binary string, and adds the bigram
to the set of vocabularies.
The above process is repeated until the number of vocabularies reaches the limit.

By this random assignment, it is expected that any sufficiently long substring contains a landmark.
Furthermore, we note that two different landmarks never overlap each other.
Therefore, LCP can merge bigrams appropriately, 
avoiding undesirable subword segmentation that occurs in BPE. 
Using this characteristics, LCP has been theoretically shown to achieve almost optimal compression~\cite{Jez2016}. 
The mechanism of LCP has also been applied to mainly information retrieval~\cite{Cormode2007,Takabatake2017,Ganczorz2018}.

A notable feature of the stochastic algorithm is that LCP assigns a new label to each vocabulary for each execution.
Owing to this randomness, the LCP-based subword segmentation is expected to generate 
different subword sequences representing a same input; 
thus, it is more robust than BPE/BPE-dropout.
Moreover, these multiple subword sequences can be considered as data augmentation for small training data in NMT.

LCP-dropout consists of two strategies: landmark by random labeling for all vocabularies
and dropout of merging bigrams depending on the rank in the frequency table.
Our algorithm requires no segmentation training in addition to counting by BPE and labeling by LCP
and uses standard BPE/LCP in test time; therefore is simple.
With various language corpora including small datasets, we show that 
LCP-dropout outperforms the baseline algorithms: BPE/BPE-dropout/SentencePiece.

%% file: sec2.tex
\section{Background}
We use the following notations throughout this paper.
Let ${\cal A}$ be the set of alphabet symbols, including the blank symbol.
A sequence $S$ formed by symbols is called a string.
$S[i]$ and $S[i,j]$ are $i$-th symbol and substring from $S[i]$ to $S[j]$ of $S$, respectively.
We assume the meta symbol `$-$' not in ${\cal A}$ to explicitly represent each subwords in $S$.
For a string $S$ from ${\cal A}\cup\{-\}$, a maximal substring of $S$ including no $-$ is called a subword.
For example, $S=a-b-a-a-b/a-b-a-ab$ contains the subwords in $\{a,b\}/\{a,b,ab\}$, respectively.

In subword segmentation, the algorithm decomposes all the symbols in $S$ by the meta symbol.
When a trigram $a-b$ is merged, the meta symbol is erased and the new subword $ab$ is added to the vocabulary, i.e.,
$ab$ is treated as a single vocabulary.

In the following, we describe previously proposed subword segmentation algorithms, called
SentencePiece (Kudo~\cite{Kudo2018}), BPE (Sennrich et al.~\cite{Sennrich2016}), 
and BPE-dropout (Provilkov et al.~\cite{Provilkov2020}), respectively.
We assume that our task in NMT is to predict a target sentence $T$ 
given a source sentence $S$, where these methods including our approach are not task-specific.

\subsection{SentencePieace}
SentencePiece~\cite{Kudo2018} can generate different segmentations for each execution.
Here, we outline SentencePiece in the unigram language model.
Given a set of vocabularies, $V$, a sentence $T$, and the probability $p(x)$ of occurrence of $x\in V$, 
the probability of the partition $x =(x_1,\ldots,x_n)$ for $T=x_1\cdots x_n$ is represented as $P(x)=\Pi_{i=1}^n p(x_i),\; x_i\in V $,
where $\Sigma_{x\in V}p(x)=1$.
The optimum partition $x^*$ for $T$ is obtained by searching for the $x$ that maximizes $P(x)$ 
from all candidate partitions $x\in S(T)$.

Given a set of sentences, $D$, as training data for a language, the subword segmentation for $D$ 
can be obtained through the maximum likelihood estimation of the following ${\cal L}$ with $P(x)$ 
as a hidden variable by using EM algorithm, where $X^{(s)}$ is the $s$-th sentence in $D$.

\[
{\cal L} = \sum_{s=1}^{|D|} \log P(X^{(s)}) = \sum_{s=1}^{|D|}\log \left( \sum_{x\in S(X^{(s)})}P(x)\right)
\]

SentencePieace was shown to achieve significant improvements over the method based on subword sequences. 
However, this method is rather complicated because it requires a unigram language model to 
predict the probability of subword occurrence, 
EM algorithm to optimize the lexicon, and Viterbi algorithm to create segmentation samples.

\subsection{BPE and BPE-dropout}
BPE~\cite{Byte-Pair-Encoding1994} is one of practical implementations of Re-pair~\cite{Re-pair2000}, 
which is known as the algorithm with the highest compression ratio.
Re-pair counts the frequency of occurrence of all bigrams $xy$ in the input string $T$.
For the most frequent $xy$, it replaces all occurrences of $xy$  in $T$ such that $T[i,i+1]=xy$, with some unused character $z$.
This process is repeated until there are no more frequent bigrams in $T$.
The compressed $T$ can be recursively decoded by the stored substitution rules $z\to xy$.

Since the naive implementation of Re-pair requires $O(|T|^2)$ time, we use a complex data structure to achieve linear-time.
However, it is not practical for large-scale data because it consumes $\Omega(|T|)$ of space.
Therefore, we usually split $T=t_1t_2\cdots t_m$ into substrings of constant length and process each $t_i$ 
by the naive Re-pair without special data structure, called BPE.
Naturally, there is a trade-off between the size of the split and the compression ratio.
BPE-based subword segmentation~\cite{Sennrich2016} (called BPE simply) 
determines the priority of bigrams according to their frequency and adds the merged bigrams as the vocabularies.

Since BPE is a deterministic algorithm, it splits a given $T$ in one way.
Thus, it is not easy to generate multiple partitions like stochastic approach (e.g.,~\cite{Kudo2018}).
Therefore, BPE-dropout~\cite{Provilkov2020}, ignoring the merging process with a certain probability, has been proposed.
In BPE-dropout, for the current $T$ and the most frequent $xy$, for each occurrence $i$ satisfying $T[i,i+1]=xy$, 
merging $xy$ is dropped with a certain small probability $p$ (e.g., $p=0.1$).
This mechanism makes BPE-dropout probabilistic and generates a variety of splits.
BPE-dropout has been recorded to outperform SentencePieace in various languages.
Additionally, BPE-based methods are faster and easier to implement than likelihood-based approaches.

\subsection{LCP}
Frequency-based compression algorithms (e.g.~\cite{Byte-Pair-Encoding1994,Re-pair2000}) are 
known to be not optimum in theoretical point of view.
Optimum compression here means a polynomial-time algorithm that satisfies 
$|A(T)| = O(|A^*(T)|\cdot \log |T| )$ with the output $A(T)$ of the algorithm for the input $T$ 
and an optimum solution $A^*(T)$.
Note that computing $A^*(T)$ is NP-hard~\cite{Lehman2002-1}.

For example, consider a string $T=\cdots \text{abcdefg}\cdots \text{bcdefg}\cdots$.
Assuming the rank of these frequencies: $freq(\text{ab})>freq(\text{bc})>freq(\text{cd})>\cdots $, 
merging for $T$ is possibly $T=\cdots \text{(ab)(cd)(ef)(g}\cdots \text{(bc)(de)(fg)}\cdots$.
However, the desirable merging would be
$T=\cdots \text{a(bc)(de)(fg)}\cdots \text{(bc)(de)(fg)}\cdots$ considering the similarity of these substrings.

Since such pathological merging cannot be prevented by frequency information alone, 
frequency-based algorithms cannot obtain asymptotically optimum compression~\cite{Lehman2002-2}.
Various linear-time and optimal compressions have been proposed to improve this drawback.
LCP is one of the simplest optimum compression algorithms.
The original LCP, like Re-pair, is a deterministic algorithm.
Recently, the introduction of probability into LCP~\cite{Jez2016} has been proposed, 
and in this study, we focus on the probabilistic variant.
The following is a brief description of the probabilistic LCP.

We are given an input string $T=a_1a_2\cdots a_n$ of length $n$ and a set of vocabularies, $V$.
Here, $V$ is initialized as the set of all characters appearing in $T$.
\begin{enumerate}
\item Randomly assign a label $L(a)\in \{0,1\}$ to each $a\in V$.
\item According to $L(a)$, compute the sequence $L(T)=L(a_1)L(a_2)\cdots L(a_n)\in \{0,1\}^n$.
\item Merge all bigram $a_ia_{i+1}$ provided $L(T)[i,i+1]=\text{`10'}$. 
\item Set $V= V\cup \{ a_ia_{i+1} \}$ and repeat the above process.
\end{enumerate}

The difference between LCP and BPE is that 
BPE merges bigrams with respect to frequencies, whereas LCP pays no attention to them.
Instead, LCP merges based on the binary labels assigned randomly.
The most important point is that any two occurrences of `10' never overlap.
For example, when $T$ contains a trigram $abc$, there is no possible assignment 
allowing $(ab)c$ and $a(bc)$ simultaneously.
By this property, LCP can avoid the problem that frequently occurs in BPE.
Although LCP theoretically guarantees almost optimum compression, 
as for as the authors know, this study is the first result of applying LCP to machine translation.

%% file: sec3.tex
\section{Our Approach: LCP-dropout}

BPE-dropout allows diverse subword segmentation for BPE by ignoring bigram 
merging with a certain probability.
However, since BPE is a deterministic algorithm, it is not trivial to generate various candidates of bigram.
In this study, we propose an algorithm that enables multiple subword segmentation 
for the same input by combining the theory of LCP with the original strategy of BPE.

\subsection{Algorithm description}
We define the notations used in our algorithm.
Let ${\cal A}$ be an alphabet and `$\_$' be the explicit blank symbol not in ${\cal A}$.
A string $w$ formed from ${\cal A}$ is called {\em word}, denoted by $x\in{\cal A}^*$,
and a string $s\in ({\cal A}\cup \{\_\})^*$ is called a {\em sentence}.

We also assume the meta symbol `$-$' not in ${\cal A}\cup\{\_\}$.
By this, a sentence $x$ is extended to have all possible merges:
Let $\tilde{x}$ be the string of all symbols in $x$ separated by $-$,
e.g., $\tilde{x} = a-b-a-b-b$ for $x=ababb$.
For strings $x$ and $y$, if $y$ is obtained by removing some occurrences of $-$ in $x$,
then we express the relation $y\preceq x$ and 
$y$ is said to be a {\em subword segmentation } of $x$.

After merging $a-b$ (i.e., $a-b$ is replaced by $ab$),
the substring $ab$ is treated as a single symbol.
Thus, we extend the notion of bigram to vocabularies of length more than two.
For a string of the form $s=\alpha_1 - \alpha_2 - \cdots -\alpha_n$ such that each $\alpha_i$ contains no $-$,
each $\alpha_i - \alpha_{i+1}$ is defined to be a bigram consisting of 
the vocabularies $\alpha_i$ and $\alpha_{i+1}$.

\renewcommand{\algorithmicrequire}{\textbf{Input:}}
\renewcommand{\algorithmicensure}{\textbf{Output:}}

\begin{algorithm}   
\caption{ \quad \bf LCP-dropout}                    
\label{lcp}                          
\begin{algorithmic}[1] 
\REQUIRE  $\tilde{X}=\{\tilde{x}_1, \tilde{x}_2, \cdots, \tilde{x}_n\}$
for a set of sentences, $X=\{x_1,x_2,\ldots, x_n\}$,
and hyperparameters $(v,\ell,k)$
\COMMENT{$v>0$: \#total vocabularies, $0<\ell \leq v$: \#partial vocabularies, $k\in (0,1]$: threshold of frequencies}

\ENSURE Set of subword sequences, ${\cal Y} = \{Y_1,Y_2,\ldots,Y_m\}$, where
$Y_i = (y_1^{(i)},y_2^{(i)},\ldots,y_n^{(i)})$ satisfies $y_j^{(i)}\preceq x_j$, $|V| = |\bigcup_{1\leq i\leq m} V(Y_i)|\leq v$ and $|V_i|=|V(Y_i)|\leq \ell$ 

\STATE $m\gets 1$ and $Y_m\gets \tilde{X}$
\WHILE{(TRUE)}
\STATE initialize $V_m,FREQ(k)$ 
\WHILE{($|V_m|<\ell$)} 
\STATE $LCP(Y_m,V_m,FREQ(k))$
\ENDWHILE
\IF{($|V|\geq v$)} 
\RETURN ${\cal Y}=\{Y_1,\ldots, Y_m\}$
\ENDIF
\STATE $m\gets m+1$, $Y_m\gets \tilde{X}$
\ENDWHILE

\end{algorithmic}
\end{algorithm}


\begin{algorithm}   
\caption{ \quad $LCP(Y,V,FREQ(k))$ \qquad \%subroutine of LCP-dropout}                    
\label{bpe}                          
\begin{algorithmic}[1] 
\STATE assign $L:V\to\{0,1\}$ randomly
\STATE $FREQ(k)\gets$ the set of top-$k$ frequent bigrams in $Y$ of the form $\alpha-\beta$ with
$L(\alpha\beta)=\text{`10'}$
\STATE merge all occurrences of $\alpha-\beta$ in $Y$ for each $\alpha-\beta \in FREQ(k)$
\STATE add all the vocabularies $\alpha\beta$ to $V$
\end{algorithmic}
\end{algorithm}

\subsection{Example run}

\begin{table}[t]
\begin{center}
\caption{
Example of multiple subword segmentation using LCP-dropout for the single sentence `$x=ababcaacabcb$'
with the hyperparameters $(v,\ell,k)=(6,5,0.5)$, where
the meta symbol $-$ is omitted, and $L$ is the label of each vocabulary assigned by LCP.
The resulting subword segmentation is ${\cal Y}=\{Y_1,Y_2\}$.
}
\label{tab:lcp}
\begin{tabular}[t]{rlllllllllllll}\hline
{\bf 1-st trial} &&  &  &  &  &  &  &  &  &  &  &  &  \\ \hline 
input $x$: depth $0$ && $a$ & $b$ & $a$ & $b$ & $c$ & $a$ & $a$ & $c$ & $a$ & $b$ & $c$ & $b$ \\ 
$L$ && $1$ & $0$ & $1$ & $0$ & $0$ & $1$ & $1$ & $0$ & $1$ & $0$ & $0$ & $0$ \\ \hline

depth $1$ && $ab$ &  & $ab$ &  & $c$ & $a$ & $a$ & $c$ & $ab$ &  & $c$ & $b$ \\ 
$L$ && $1$ &  & $1$ &  & $0$ & $1$ & $1$ & $0$ & $1$ &  & $0$ & $1$ \\ \hline

$Y_1$: depth $2$ && $ab$ &  & $abc$ &  &  & $a$ & $a$ & $c$ & $abc$ &  & & $b$ \\ \hline \hline

{\bf 2-nd trial} &&  &  &  &  &  &  &  &  &  &  &  &  \\ \hline

same $x$: depth $0$  &&   $a$ & $b$ & $a$ & $b$ & $c$ & $a$ & $a$ & $c$ & $a$ & $b$ & $c$ & $b$ \\  
$L$ && $0$ & $1$ & $0$ & $1$ & $1$ & $0$ & $0$ & $1$ & $0$ & $1$ & $1$ & $1$ \\ \hline

depth $1$ && $a$ & $b$ & $a$ & $b$ & $ca$ &  & $a$ & $ca$ &  & $b$ & $c$ & $b$ \\ 
$L$ &&                    $1$ & $0$ & $1$ & $0$ & $0$  &      &  $1$ & $0$ &  & $0$ & $0$ & $0$ \\ \hline

$Y_2$: depth $2$ && $ab$ &  & $ab$ &  & $ca$ &  & $a$ & $ca$ &  & $b$ & $c$ & $b$ \\ \hline
\end{tabular}
\end{center}
\end{table}

Table~\ref{tab:lcp} presents an example of subword segmentation using LCP-dropout.
Here, the input $X$ consists of a single sentence $ababcaacabcb$.
The hyperparameters are $(v,\ell,k)=(6,5,0.5)$.
First, the set of vocabularies is initialized to $V=\{a,b,c\}$; for each $\alpha\in V$,
a label $L(w)\in\{0,1\}$ is randomly assigned (depth 0).
Next, find all occurrences of $10$ in $L$, and the corresponding bigrams are merged depending on their frequencies.
Here, $L(ab)=L(ac)=10$ but only $a-b$ is top-$k$ bigram assigned $10$, and then $a-b$ is merged to $ab$.
The resulting string is shown in the depth $1$ over the new vocabularies $V_1=\{a,b,c,ab\}$.
This process is repeated while $|V_m|<\ell$ for the next $m$.
The condition $|V_2|=5$ terminates the inner-loop of LCP-dropout, and then
the subword $Y_1 = ab-abc-a-a-c-abc-b$ is generated.
Since $|V(Y_1)|<v$, the algorithm generates the next subword segmentations $Y_2$ for the same input.
Finally, we obtain the multiple subword segmentation 
$Y_1=ab-abc-a-a-c-abc-b$ and $Y_2=ab-ab-ca-a-ca-b-c-b$
for the same input string. 

\subsection{Framework of neural machine translation}

\begin{figure}[t]
    \begin{center}
    \includegraphics[width=10cm]{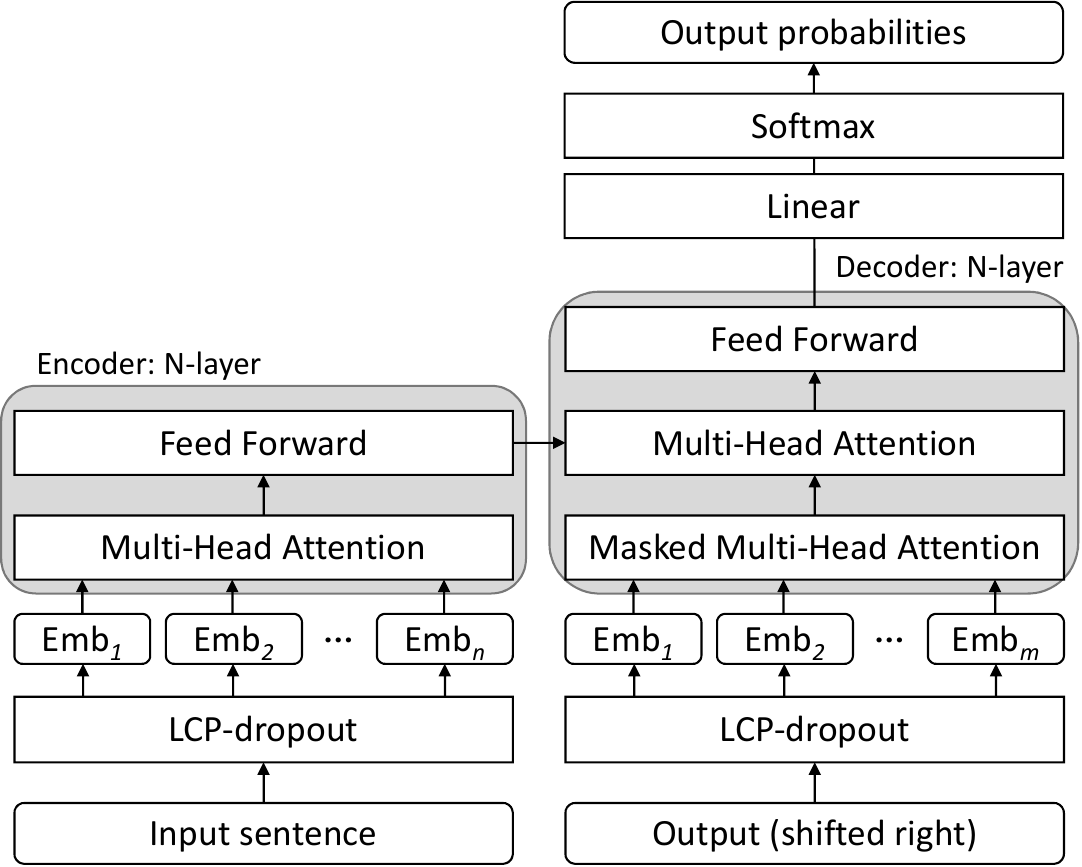}
    \caption{Framework of our neural machine translation model with LCP-dropout.}\label{fig:nmt}
    \end{center}
\end{figure}

Figure \ref{fig:nmt} shows the framework of our transformer-based machine translation model with LCP-dropout.
Transformer is the most successful NMT model \cite{Vaswani2017}.
The model mainly consists of Encoder and Decoder. 
The Encoder converts the input sentence in the source language into a word embedding (Emb$_i$ in Figure \ref{fig:nmt}), taking into account the positional information of the characters. 
Here, the notion of word is extended to that of subword in this study. 
The subwords are obtained by our proposed method, LCP-dropout.
Next, the correspondences in the input sentence are acquired as attention (Multi-Head Attention).
Then, the normalization is done through a forward propagation network formed by linear transformation, activation by ReLU function, and linear transformation. 
These processes are performed in N = 6 layers for the Decoder.

For the Decoder, it receives the candidate sentence generated by the Encoder and the input sentence for the Decoder.
Then, it acquires the correspondence between those sentences as attention (Multi-Head Attention). 
This process is also performed in N = 6 layers.
Finally, the predicted probability of each label is calculated by linear transformation and softmax function.

%% file: sec4.tex
\section{Experimental Setup}

\subsection{Baseline algorithms}
Baseline algorithms are SentencePiece~\cite{Kudo2018} with the unigram language model
and BPE/BPE-dropout~\cite{Sennrich2016,Provilkov2020}.
SentencePiece takes the hyperparameters $l$ and $\alpha$, where
$l$ specifies how many best segmentations for each word are produced before sampling
and $\alpha$ controls the smoothness of the sampling distribution.
In our experiment, we use $(l=64,\alpha=0.1)$ that performed best on different data in the previous studies.

BPE-dropout takes the hyperparameter $p$, where during segmentation, at each step, some merges are 
randomly dropped with the probability $p$.
If $p=0$, the segmentation is equal to the original BPE and $p=1$, the algorithm outputs the input string itself.
Then, the value of $p$ can be used to control the granularity of segmentation.
In our experiment, we use $p=0$ for the original BPE and $p=0.1$ for the BPE-dropout with the best performance.

\subsection{Data sets, preprocessing, and vocabulary size}
\begin{table}[t]
\begin{center}
\caption{
Overview of the datasets and hyperparameters.
The hyperparameter $v$ (vocabulary size) is common to all algorithms (baselines and ours)
and others ($\ell$ and $k$) are specific to LCP-dropout only.
}
\label{tab:data}
\begin{tabular}[t]{ccccc}\hline
corpus & language & \#sentences & batch & hyperparameters \\
& ($L_1- L_2$) & (train/dev/test) & size & $(v,\ell,k)$ \\
\hline
News & {\bf En} $-$ {\bf De} & 380k/2808/2906 & 3072 & 16k, 16k/8k, 0.01/0.05/0.1 \\
Commentary & {\bf En} $-$ {\bf Fr} & 357k/3020/3133 & 3072 & 16k, 8k, 0.01 \\
v16& {\bf En} $-$ {\bf Zh} & 305k/2968/2936 & 3072 & 16k, 8k, 0.01 \\
\hline
KFTT & {\bf En} $-$ {\bf Ja} & 440k/1166/1160 & 3072 & 16k, 8k, 0.01 \\
\hline
WMT14 & {\bf En} $-$ {\bf De} & 4.5M/2737/3004 & 3072 & 32k, 32k/16k, 0.01 \\
\hline
\end{tabular}
\end{center}
\end{table}

We verify the performance of the proposed algorithm for a wide range of datasets with different sizes and languages.
Table~\ref{tab:data} summarizes the details of the datasets and hyperparameters.
These data are used to compare the performance of LCP-dropout
and baselines (SentencePiece/BPE/BPE-dropout) with appropriate hyperparameters
and vocabulary sizes shown in~\cite{Provilkov2020}.

Before subword segmentation, we preprocess all datasets with the standard Moses toolkit,\footnote{https://github.com/moses-smt/mosesdecoder}
where for Japanese and Chinese, subword segmentations are trained almost from raw sentences
because these languages have no explicit word boundaries; and thus, Moses tokenizer does not work correctly.

Based on a recent research on the effect of vocabulary size on translation quality, 
the vocabulary size is modified according to the dataset size in our experiments (Table~\ref{tab:data}).

To verify the performance of the proposed algorithm for small training data,
we use News Commentary v16\footnote{https://data.statmt.org/news-commentary/v16}, a subset of 
WMT14\footnote{https://www.statmt.org/wmt14/translation-task.html}, as well as KFTT\footnote{http://www.phontron.com/kftt}.
In addition, we use a large training data in WMT14.
The training step is set to 200,000 for all data.
In training, pairs of sentences of source and target languages were batched together by approximate length.
As shown in Table~\ref{tab:data}, the batch size was standardized to approximately $3\text{k}$ for all datasets.

\subsection{Model, optimizer, and evaluation}
NMT was realized by the {\em seq2seq} model, which takes a sentence in the source language as input and 
outputs a corresponding sentence in the target language~\cite{Sutskever2014}.
A transformer is an improvement of seq2seq model, that is the most successful NMT.~\cite{Vaswani2017}

In our experiments, we used OpenNMT-tf~\cite{Klein2017}, 
a transformer-based NMT, to compare LCP-dropout and other baselines algorithms.
The parameters of OpenNMT-tf were set as in the experiment of BPE-dropout~\cite{Provilkov2020}.
The batch size was set to 3072 for training and 32 for testing.
We also use regularization and optimization procedure as described in BPE-dropout~\cite{Provilkov2020}.

The quality of machine translation is quantitatively evaluated by BLEU score,
i.e., the similarity between the result and the reference of translation.
It is calculated using the following formula based on the number of matches in their $n$-grams.
Let $t_i$ and $r_i$ $(1\leq i\leq m)$ be the $i$-th translation and reference sentences, respectively.

\[
\text{BLEU} = BP_{\text{BLEU}}\cdot \exp\left(\sum_{n=1}^N w_n\log p_n \right),\;\; p_n = \frac{\sum_{i=1}^m \#n\text{-gram that match in $t_i$ and $r_i$}}
{\sum_{i=1}^m \#n\text{-gram in $t_i$}},
\]

\noindent
where $N$ is a small constant (e.g., $N=4$), and 
$BP_{\text{BLEU}}$ is the brevity penalty when $|t_i|<|r_i|$, where
$BP_{\text{BLEU}}=1$ otherwise.
In this study, we use SacreBLEU~\cite{Post2018}. 
For Chinese, we add option{\tt  --tok zh} to SacreBLEU. 
Meanwhile, we use character-based BLEU for Japanese.

%% file: sec5.tex
\section{Experiments and Analysis}

All experiments are conducted in the environments:
OS: Ubuntu 20.04.2 LTS, 
CPU: Intel(R) Xeon(R) W-2135 CPU @ 3.70GHz,
GPU: GeForce RTX 2080 Ti Rev. A,
Memory: 64GB RAM,
Storage: 2TB SSD, 
Python-3.8.6,
SentencePiece-0.1.96~\footnote{\tt https://pypi.org/project/sentencepiece/} 
(Python wrapper for SentencePiece including BPE/BPE-dropout runtime).
The numerical results are averages of three independent trials.

\subsection{Estimation of hyperparameters for LCP-dropout}

\begin{table}[t]
\begin{center}
\caption{
Experimental results of LCP-dropout for News Commentary v16 (Table~\ref{tab:data}) 
w.r.t the specified hyperparameters, where the translation task is {\bf De}$\to${\bf En}.
Bold indicates the best score.
}
\label{tab:result1-1}
\begin{tabular}[t]{lllllll}\hline
top-$k$ threshold && \#subword &&& BLEU &\\
$(k\in (0,1])$ && {\bf En} & {\bf De} & &$(\ell = v)$ & $(\ell = v/2)$ \\ \hline
0.01 && 21.3 & 7.7 &&  39.0 & {\bf 39.7}\\
0.05 && 4.7 & 3.7 & & 39.0 & {\bf 39.4} \\
0.1 && 3.3 & 2.0 & & 38.8  & {\bf 39.4} \\ \hline
\end{tabular}
\end{center}
\end{table}

First, we estimate suitable hyperparameters for LCP-dropout.
Table~\ref{tab:result1-1} summarizes the effect of hyperparameters $(v,\ell,k)$ on the proposed LCP-dropout.
This table shows the details of multiple subword segmentation using LCP-dropout and BLEU scores for 
the language pair of English ({\bf En}) and German ({\bf De}) from News Commentary v16 (Table~\ref{tab:data}).
For each threshold $k\in\{0.01,0.05,0.1\}$, {\bf En} and {\bf De} indicate the number of multiple subword sequences generated 
for the corresponding language, respectively.
The last two values are the BLEU scores for {\bf De} $\to$ {\bf En}
with $\ell = v$ and $\ell=v/2$ for $v=16\text{k}$, respectively.

\begin{table}[t]
\begin{center}
\caption{
Depth of label assignment in LCP-dropput.
}
\label{tab:result1-2}
\begin{tabular}[t]{llllll}\hline
top-$k$ threshold & $\ell = v$ && & $\ell = v/2$ &  \\ 
$(k\in (0,1])$ & {\bf En} & {\bf De} && {\bf En} & {\bf De} \\ \hline
0.01 & 83.7 & 48.0 && 54.9 & 35.4 \\
0.05 & 18.7 & 12.3 && 13.0 & 9.0 \\
0.1 & 10.3  & 7.7 && 7.3 & 6.0 \\ \hline
\end{tabular}
\end{center}
\end{table}

The threshold $k$ controls the dropout rate, and $\ell$ contributes to the multiplicity of the subword segmentation.
The results show that $k$ and $\ell$ affect the learning accuracy (BLEU). 
The best result is obtained when $(k,\ell)=(0.01,\ell=v/2)$.
This can be explained by the results in Table~\ref{tab:result1-2} which shows
the depth of executed inner-loop of LCP-dropout for randomly assigning $\{0,1\}$ to vocabularies,
where when $\ell =v/2$, it means the average before the outer-loop terminates.
Therefore, the larger this value is, the more likely it is that longer subwords will be generated.
However, unlike BPE-dropout, the value of $k$ alone is not enough to generate multiple subwords.
The proposed LCP-dropout guarantees the diversity by initializing the subword segmentation by $\ell$ $(\ell <v)$.
Using this result, we will fix $(k,\ell)=(0.01,v/2)$ as the hyperharameter of LCP-dropout.

\subsection{Comparison with baselines}

\begin{table}[t]
\begin{center}
\caption{
Experimental results of LCP-dropout (denoted by LCP), BPE-dropout (denoted by BPE), and SentencePiece
(denoted by SP)
 on various languages in Table~\ref{tab:data} (small corpus: News Commentary v16 and 
 KFTT, and large corpus: WMT14), 
wehre `multiplicity' denotes the average number of sequences generated per input string.
Bold indicates the best score.
}
\label{tab:result1-3}
\begin{tabular}[t]{lllllll}\hline
Corpus&Language & Translation & LCP & BPE &  & SP \\ 
& (multiplicity)& Direction&$(k=0.01)$&$(p=1,$&$0.1)$& \\ \hline
News& {\bf En}$-${\bf De}  & {\bf De} $\to$ {\bf En} & {\bf 39.7} &  35.7 & 39.1   & 38.9 \\
Commentary& (21.3-7.7)& {\bf En} $\to$ {\bf De} & {\bf 28.4} & 27.4 & 27.4 & 27.5 \\ 
v16& {\bf En}$-${\bf Fr}   & {\bf Fr} $\to$ {\bf En} & {\bf 35.1} &  34.9 &34.9  & 34.2 \\
(small)& (23.0-19.3)& {\bf En} $\to$ {\bf Fr} & {\bf 29.5} & $15.2$ & 28.2 & 28.3 \\ 
& {\bf En}$-${\bf Zh}   & {\bf Zh} $\to$ {\bf En} &24.2 & 24.2 & {\bf 24.6} & 24.2 \\
& (26.0-8.7)& {\bf En} $\to$ {\bf Zh} & {\bf 6.5}  & $2.0$ & 2.1 & $1.8$ \\ \hline 
KFTT& {\bf En}$-${\bf Ja}   & {\bf Ja} $\to$ {\bf En} & {\bf 20.0} & 19.6 & 19.6 & 19.2 \\
(small)&  (17.7-10.0)& {\bf En} $\to$ {\bf Ja} & {\bf 8.5} & $3.0$ & 3.6 & $3.5$ \\ \hline
WMT14& {\bf En}$-${\bf De}   & {\bf De} $\to$ {\bf En} &28.7&28.9&{\bf 32.2}&{\bf 32.2} \\ 
(large)& (9.3-5.3) &&&&& \\ \hline
\end{tabular}
\end{center}
\end{table}

\begin{table}[t]
\begin{center}
\caption{
Depth of label assignment for large corpus.
}
\label{tab:result1-4}
\begin{tabular}[t]{llllll}\hline
top-$k$ threshold & $\ell = v$ && & $\ell = v/2$ &  \\ 
$(k\in (0,1])$ & {\bf En} & {\bf De} && {\bf En} & {\bf De} \\ \hline
0.01 & 24.0 & 18.0 && 17.1 & 14.2 \\ \hline
\end{tabular}
\end{center}
\end{table}

Table~\ref{tab:result1-3} summarizes the main results.
We show BLEU scores for News Commentary v16 and KFTT:
{\bf En} and {\bf De} are the same in Table~\ref{tab:result1-1}.
In addition to these languages, we set French ({\bf Fr}), Japanese ({\bf Ja}), and Chinese ({\bf Zh}).
For each language, we show the average of the number of multiple subword sequences generated by LCP-dropout.
For almost datasets, LCP-dropout outperforms the baselines algorithms.
Meanwhile, we use the best ones reported in the previous study for the hyperparameters of 
BPE-dropout and SentencePiece.

Table~\ref{tab:result1-3} extracts the effect of alphabet size on subword segmentation.
In general, Japanese {\bf Jn} and Chinese {\bf Zh} alphabets are very large, 
containing at least $2\text{k}$ alphabet symbols even if we limit them in common use.
Therefore, the average length of words is small and subword semantics is difficult.
For these case, we confirmed that LCP-dropout has higher BLEU scores than other methods for these languages.

Table~\ref{tab:result1-3} also presents the BLEU scores for a large corpus (WMT14)
for the translation {\bf De} $\to$ {\bf En}.
This experiment shows that LCP-dropout cannot outperform baselines with the hyperparameter we set.
This is because the ratio of the vocabulary size $(v,\ell)$ to dropout rate $k$ is not appropriate.
As data to support this conjecture, it can be confirmed that the multiplicity in the large datasets 
is much smaller than that of small corpus (Table~\ref{tab:result1-3}).
This is caused by the reduced repetitions of label assignments, as shown in Table~\ref{tab:result1-4}
compared to Table~\ref{tab:result1-2}.
The results show that the depth of inner-loop is significantly reduced, 
which is why enough subwords sequences cannot be generated.

\begin{table}[t]
\begin{center}
\caption{
Examples of translated sentences by LCP-dropout $(k=0.01)$ and BPE-dropout $(p=0.1)$
with the reference translation for News Commentary v16.
We show the average word length (ave./word) for each reference sentence
as well as the average subword length (ave./subword) generated by respective algorithms
for the entire corpus.
We also show the BLEU sores between the references and translated sentences
as well as their standard deviations (SD).
}
\label{tab:result1-5}
\begin{tabular}[t]{rll}\hline
Reference: & `Even if his victory remains unlikely, Bayrou must  & BLEU \\ 
(ave./word = {\bf 5.00})&  now be taken seriously.' & \\ \hline
LCP-dropout: & `While his victory remains unlikely, Bayrou must& 84.5 \\
&  now be taken seriously.' & \\ 
BPE-dropout: & `Although his victory remains unlikely, he needs to& 30.8 \\ 
&  take Bayrou seriously now.' & \\ \hline \hline
Reference: & `In addition, companies will be forced to restructure & BLEU \\ 
(ave./word = {\bf 5.38})& in order to cut costs and increase competitiveness.' & \\ \hline
LCP-dropout: & `In addition, restructuring will force rms to save costs& 12.4 \\
&  and boost competitiveness.' & \\
BPE-dropout: & `In addition, businesses will be forced to restructure & 66.8 \\
& in order to save costs and increase competitiveness.' & \\ \hline \hline
ave./subword & {\bf 4.01} (LCP-dropout) : {\bf 4.31} (BPE-dropout) & \\ \hline \hline
SD of BLEU & {\bf 21.98} (LCP-dropout) : {\bf 21.59} (BPE-dropout) & \\ \hline
\end{tabular}
\end{center}
\end{table}

Table~\ref{tab:result1-5} presents several translation results.
The `Reference' represents the correct translation for each case, 
and the BLEU score is obtained from the pair of the reference and each translation result.
We also show the average length for each reference sentence indicate by `ave./word'.
These results show the characteristics of successful and unsuccessful translations by the two algorithms
related to the length of words.

Considering subword segmentation as a parsing tree, 
LCP produces a balanced parsing tree, whereas the tree produced by BPE tends to be longer for a certain path.
For example, for a substring $abcd$, 
LCP tends to generate subwords like $((ab)(cd))$, while BPE generates them like $(((ab)c)d)$.
In this example, the average length of the former is shorter than that of the latter.
This trend is supported by the experimental results in Table~\ref{tab:result1-5}
showing the average length of all subwords generated by LCP/BPE-dropout for real datasets.
Due to this property, when the vocabulary size is fixed, 
LCP tends not to generate subwords of approximate length
because it decomposes a longer word into excessively short subwords.

\begin{figure}[t]
    \begin{center}
    \includegraphics[width=10cm]{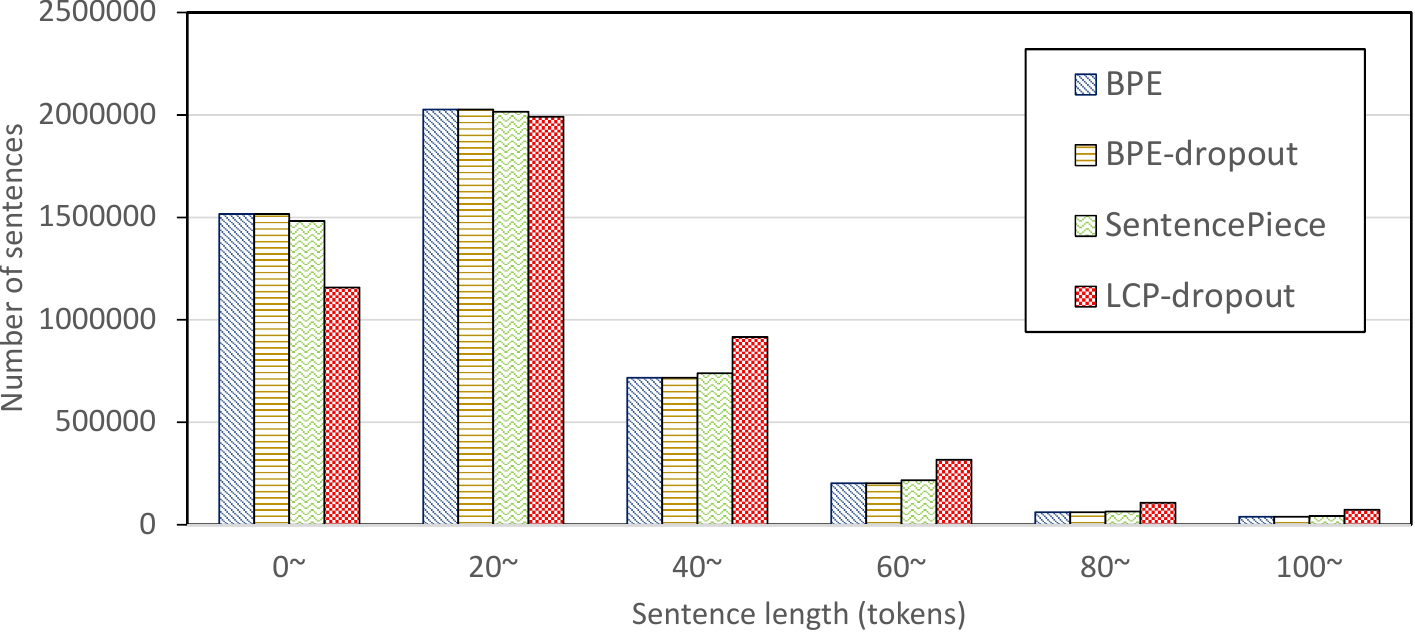}
    \caption{Distribution of sentence length. The number of tokens in each sentence by LCP-dropout tends to be larger than the others: BPE, BPE-dropout, and SentencePiece.}
    \label{fig:distribution}
    \end{center}
\end{figure}
Figure \ref{fig:distribution} shows the distributions of sentence length of English.
The sentence length denotes the number of tokens in a sentence.
BPE-dropout is a well-known fine-grained segmentation approach.
The figure shows that LCP-dropout produces more fine-grained segmentation than the other three segmentation approaches.

Therefore, LCP-dropout is considered to be superior in subword segmentation for 
languages consisting of short words.
Table~\ref{tab:result1-3} including the translation results for Japanese and Chinese
also supports this characteristics.

%% file: sec6.tex
\section{Conclusions, Limitations, and Future Research}

\subsection{Conclusions and limitations}
In this study, proposed LCP-dropout as an extension of BPE-dropout~\cite{Provilkov2020} 
for multiple subword segmentation by applying a near-optimum compression algorithm.
The proposed LCP-dropout can properly decompose strings without background knowledge 
of the source/target language by randomly assigning binary labels to vocabularies.
This mechanism allows generating consistent multiple segmentations for the same string.
As shown in the experimental results,
LCP-dropout enables data augmentation for small datasets, 
where sufficient training data are unavailable on minor languages or limited fields.

Multiple segmentation can also be achieved by likelihood-based methods.
After SentencePiece~\cite{Kudo2018}, various extensions have been proposed~\cite{Haffari2020,Deguchi2020}.
In contrast to these studies, our approach focuses on a simple linear-time compression algorithm.
Our algorithm does not require any background knowledge of the language compared to word replacement-based 
data augmentation,~\cite{Fadaee2017,Wang2018} where some words 
in the source/target sentence are swapped with other words 
preserving grammatical/semantic correctness.

\subsection{Future research}
The effectiveness of LCP-dropout was confirmed for almost small corpora.
Unfortunately, the optimal hyperparameter obtained in this study did not work well for a large corpus.
Besides, the learning accuracy was found to be affected by the alphabet size of the language.
Future research directions include an adaptive mechanism for determining the hyperparameters 
depending on training data and alphabet size.

In the experiments in this paper, we considered word-by-word subword decomposition.
On the other hand, multi-words are known to violate the compositeness of language. 
Therefore, by considering multi-words as longer words and performing subword decomposition,
LCP-dropout can be applied to related to language processing related to multi-words.
In this study, subword segmentation was applied to machine translation.
To improve the BLEU score, there are other approaches such as data augmentation~\cite{Sawai2022}.
Incorporating the LCP-dropout with them is one interesting approach.
In this paper, we handled several benchmark datasets with major languages.
Recently, machine translation of low-resource languages is an important task~\cite{Park2020}.
Applying the LCP-dropout to the task is also important future work.

Although proposed LCP-dropout is currently applied only to machine translation, 
we plan to apply our method to other linguistic tasks 
including sentiment analysis, parsing, and question answering in future studies.

